\documentclass{article}

\usepackage{arxiv}
\usepackage{graphicx}
\usepackage[utf8]{inputenc} % allow utf-8 input
\usepackage[T1]{fontenc}    % use 8-bit T1 fonts
\usepackage{algorithmic}
\usepackage[ruled,vlined,linesnumbered]{algorithm2e}
\usepackage{multirow}
\usepackage{threeparttable}
\usepackage{amsmath}
\usepackage{amssymb}
\usepackage{amsfonts}

\title{A dynamic memory assignment strategy for dilation-based ICP algorithm on embedded GPUs}

\author{
 Qiong Chang \\
  School of Coumputing\\
  Institute of Science Tokyo\\
  Tokyo, Japan 152-8550 \\
  \texttt{q.chang@c.titech.ac.jp} \\
   \And
 Weimin Wang\\
  Dalian University of Technology\\
  Dalian, China 210093  \\
  \texttt{wangweimin@dlut.edu.cn} \\
  \And
 Junpei Zhong\\
  University of Wollongong (College HK)\\
  Hong Kong, China  \\
  \texttt{joni.zhong@ieee.org} \\
  \And
 Jun Miyazaki \\
  School of Computing\\
  Institute of Science Tokyo\\
  Tokyo, Japan 152-8550 \\
  \texttt{miyazaki@c.titech.ac.jp} \\
}

\begin{document}
\maketitle
\begin{abstract}
This paper proposes a memory-efficient optimization strategy for the
high-performance point cloud registration algorithm VANICP, enabling lightweight
execution on embedded GPUs with constrained hardware resources. VANICP is a
recently published acceleration framework that significantly improves the
computational efficiency of point-cloud-based applications. By transforming the
global nearest neighbor search into a localized process through a dilation-based
information propagation mechanism, VANICP greatly reduces the computational
complexity of the NNS. However, its original implementation demands a
considerable amount of memory, which restricts its deployment in
resource-constrained environments such as embedded systems. To address this
issue, we propose a GPU-oriented dynamic memory assignment strategy that
optimizes the memory usage of the dilation operation. Furthermore, based on this
strategy, we construct an enhanced version of the VANICP framework that achieves
over 97\% reduction in memory consumption while preserving the original
performance. Source code is published on:
\textit{https://github.com/changqiong/VANICP4Em.git}.
\end{abstract}

% keywords can be removed
\section{Introduction}
The Iterative Closest Point (ICP) algorithm is one of the most fundamental
methods in 3D vision, widely used for point cloud registration in tasks such as
tracking~\cite{roussel2021unsupervised}, and object
recognition~\cite{maken2021stein,vizzo2023kiss}. However, ICP is computationally
intensive, as its performance is dominated by the time-consuming nearest
neighbor search (NNS) process~\cite{besl1992method}. Recently, many studies have
focused on accelerating point cloud registration to enhance its applicability in
real-world
scenarios~\cite{koide2021voxelized,koide2021globally}. Alexander~\cite{agathos2024multi}
proposed a multi-GPU strategy that efficiently performances nearest neighbor
search on large-scale point clouds. Yu et al.~\cite{yu2025fsac} employed a
Hierarchical Navigable Small World (HNSW) structure to accelerate descriptor
matching within their point cloud alignment framework. More recently,
VANICP~\cite{chang2025accelerating} demonstrates significant performance gains
by transforming the global NNS into a localized process through a dilation-based
strategy. It completes the registration of the {\em Standford bunny} dataset
(containing over 40k points) within 0.5s on an RTX 3080 GPU. However, it still
suffers from a critical limitation that requiring preallocating a large and
fixed memory space (approximately 2GB) to support voxel dilation operations
across regions with varying sparsity. This leads to two major challenges: 1),
the substantial memory consumption renders the framework unsuitable for embedded
systems, where computational resources, energy budgets, and mobility are tightly
constrained; 2), the fixed-size memory allocation lacks adaptability, which may
result in either significant memory waste or insufficient capacity under
different datasets.
\begin{figure}[t]
\centering
\includegraphics[width=5in]{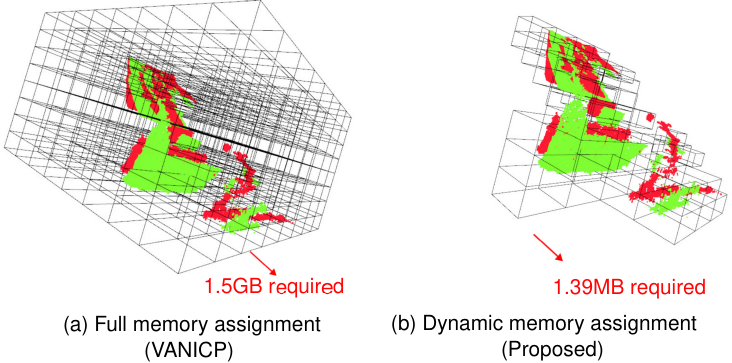}
\caption{Comparison of memory consumption in dilation. For the {\em TUM} model
  (226K points), the original VANICP adopts a monolithic contiguous assignment
  strategy for dilation, which supports direct addressing. In contrast, the
  proposed method employs a segmented pointer-based assignment strategy that
  relies on indirect addressing.}
\label{fig1}
\end{figure}

To overcome the above challenges, we proposed a GPU-oriented dynamic memory
assignment strategy tailored for embedded GPU platforms. Leveraging the unified
memory of embedded GPUs, the proposed strategy dynamically allocates memory for
each voxel in a heterogeneous manner based on the number of points contained
within it. Specifically, the method performs parallel counting of points in each
voxels, followed by a serial computation of memory start addresses to ensure
correct allocation boundaries. Experimental results demonstrate that the
proposed strategy reduces memory consumption by more than 97\%, while
maintaining the original computational performance of VANICP.

The remainder of this paper is organized as follows. Section II introduces the
original VANICP algorithm and outlines its core principles. Section III
presents the proposed GPU-oriented dynamic memory assignment strategy and its
implementation details. Section IV describes the experimental setup and
discusses the results. Section V reviews related approaches for accelerating ICP
algorithms. Finally, Section VI concludes the paper by summarizing the key
contributions and outlining potential directions for future work.

\section{Background}
\subsection{VANICP}
The goal of the basic ICP is to align a source point cloud $P=\{p_1,
p_2,...,p_m\}$ with a target point cloud $Q=\{q_1, q_2,...,q_n\}$ by iteratively
estimating a rigid transformation between them. In each iteration,
correspondences are first established by finding the nearest neighbor $q'_i$ for
each source point $p_i$. And then, ICP estimates the optimal rotation matrix
$R\in{\mathbb{R}^{3\times{3}}}$ and the translation vector $t\in{\mathbb{R}^3}$
that minimize the mean squared error function between the corresponding pairs:
\begin{equation}
\min_{\mathbf{R},\,\mathbf{t}}
\sum_i \| \mathbf{R}\mathbf{p}_i + \mathbf{t} - \mathbf{q'}_i \|^2.
\end{equation}
As the most computationally intensive step of the ICP, the nearest neighbor
search (NNS) has attracted considerable attention from researchers seeking
performance optimization. Among the proposed acceleration methods,
VANICP~\cite{chang2025accelerating, wang2023van} distinguishes itself with a
unique dilation-based design that transforms the global NNS into a localized
process. This design enables substantially higher performance compared to
traditional kd-tree-based registration
methods~\cite{greenspan2003approximate}. In VANICP, the accelerated NNS can be
decomposed into the following phases:

\noindent\textbf{Step1: Voxelization} The target point cloud is first converts
into a discrete 3D grid structure, where the space is divided into small voxels.
Each voxel contains all points that fall within its spatial boundary,
effectively grouping nearby points together. This structured representation
facilitates efficient spatial indexing and reducing the computational cost of
subsequent nearest neighbor searching.

\noindent\textbf{Step2: Voxel Dilation} This step is applied during early
iterations when the overlap between source and target point clouds is still
limited. By dilating the occupied voxels, the algorithm effectively enlarges the
overlap region, enabling more source points to find valid correspondences within
nearby target voxels. This not only improves matching robustness in the initial
stages but also accelerates the propagation of spatial information across
adjacent regions, leading to faster and more stable convergence.

\noindent\textbf{Step3: Local Approximate Nearest Neighbor Search} After
voxelization and dilation, each source point only needs to search within its
assigned voxel or its dilated neighbors, rather than across the entire target
point cloud. If the voxel is non-empty, a local search is performed (O(n)
complexity within the voxel). If the voxel is empty, the algorithm falls back to
a global search to ensure correctness, though such cases become increasingly
rare as iterations proceed.

Finally, the nearest neighbor search of each source point is mapped to an
individual GPU thread. Leveraging voxel-based locality, these approximate
nearest neighbor searches are highly parallelizable, avoiding the global kd-tree
traversals or brute-force distance computations that dominate traditional ICP.

\subsection{Memory Inefficiency Caused by Voxel Sparsity Variations}
Since the dilation operation is performed voxel-wise in parallel, each voxel
requires an independent memory block. However, voxels often exhibit significant
sparsity because the number of points they contain varies widely. Ideally,
memory allocation should be dynamic, scaling with the actual number of points in
each voxel. In practice, however, GPUs are optimized for massive parallelism and
deterministic resource management, relying on preallocated, contiguous memory
regions to sustain high throughput and minimize synchronization
overhead. Consequently, VANICP adopts a static strategy that preallocates a 64KB
memory block per voxel. This approach not only causes substantial memory
waste since many voxels contain only a few points, but also limits adaptability
to point clouds of varying density and scale. Furthermore, the large memory
utilization makes the framework unsuitable for embedded systems, where
computational resources and energy efficiency are tightly constrained.

\begin{figure*}[t]
\centering
\includegraphics[width=6.5in]{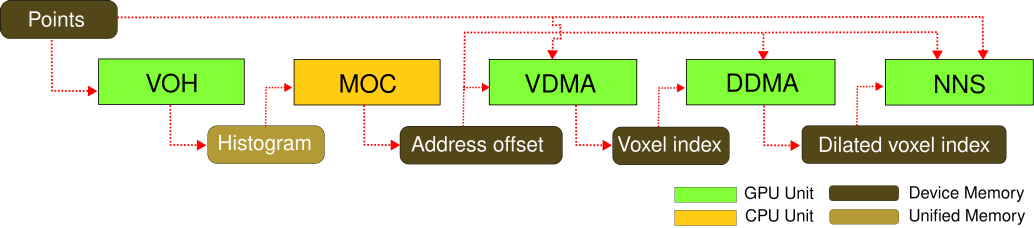}
\caption{Processing flow of the proposed framework. VOH: voxelization occupancy
  histogram. MOC: memory offset computation. VDMA: voxelization with dynamic
  memory assignment. DDMA: dilation with dynamic memory assignment. NNS: nearest
  neighbor searching. This strategy first performs voxel-wise histogram
  construction on the GPU to parallelize point counting, and then computes the
  memory offset of each voxel on the CPU via unified memory. By integrating
  offset-based indexing into subsequent processing stages, it significantly
  reduces the overall memory utilization.}
\label{fig2}
\end{figure*}

\section{Proposed Algorithm}
In this section, we plan to introduce the proposed dynamic memory assignment
strategy. 

\begin{figure}[t]
\centering
\includegraphics[width=3.5in]{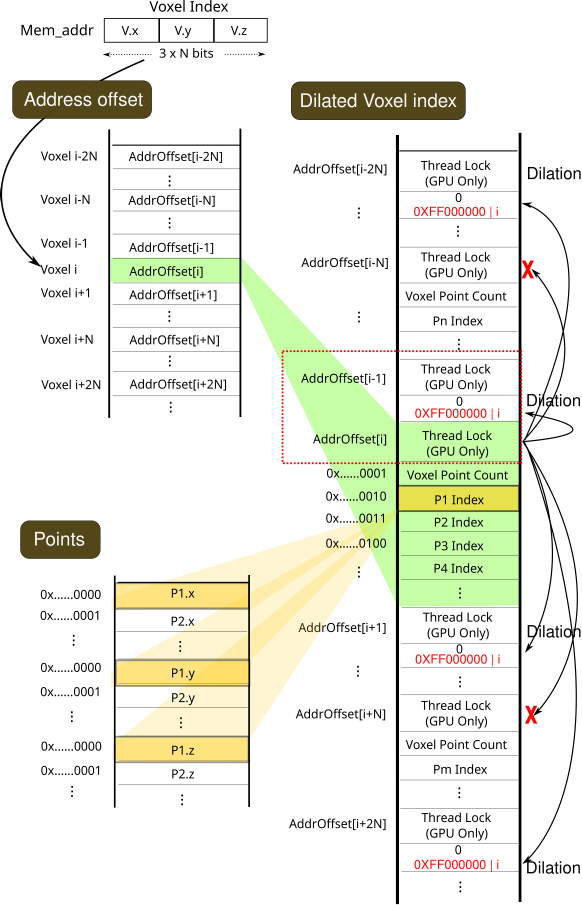}
\caption{Memory assignment of the proposed strategy.}
\label{fig3}
\end{figure}

Figure~\ref{fig2} illustrates the overall pipeline of the proposed
strategy. Unlike VANICP, which assigns a fixed memory size to each voxel, our
approach dynamically allocates memory according to the actual occupancy of the
input point cloud. In addition to the voxelization, dilation, and NNS stages
inherited from VANICP, we introduce a series of preprocessing steps that
leverage heterogeneous computation to gather voxel-wise
statistics. Specifically, a parallel histogram is constructed on the GPU to
count the number of points in each voxel, followed by a prefix-based computation
of the voxel memory offsets on the CPU. By exploiting the unified memory of
embedded GPUs, both histogram construction (parallel) and offset accumulation
(serial) can be performed efficiently within a single address space. 
The memory offset plays a crucial role in this strategy. Since it is derived
from the point count of each voxel, the offset implicitly encodes both the size
of each voxel and the total memory requirement. Instead of the original
direct-addressing strategy that relies on a monolithic memory layout, the
proposed method adopts an indirect-addressing approach through the memory
offset. Although this introduces additional memory accesses, the overhead is
outweighed by the nearly 100x reduction in memory utilization.
\begin{algorithm}
\small
\DontPrintSemicolon
\SetAlgoNlRelativeSize{-1}
\caption{Heterogeneous Address Offset Computation}
\label{alg1}

\tcp*[l]{GPU-side Voxel Occupancy Histogram Generation}
\KwIn{Target point $P_i$}
\KwOut{VoxelHist}

\ForEach{point $P_i$}{
    $V_x, V_y, V_z \gets \textnormal{Quantize}(P_i.x, P_i.y, P_i.z)$\;
    $IndexVoxel \gets \textnormal{Concat}(V_x, V_y, V_z)$\;
    $key \gets IndexVoxel \& (\textnormal{BLOCK\_SIZE} - 1)$\;
    \While{True}{
    $prev \gets \textnormal{atomicCAS}(SMem[key], 0, IndexVoxel<<16)$\;
    \If{$(prev == 0)\ \lor\ \big((prev \gg 16) == IndexVoxel\big)$}{
        $\textnormal{atomicAdd}(\&SMem[key], 1)$\;
        $\textnormal{break}$\;
    }
    $key \gets (key + 1) \& (\textnormal{BLOCK\_SIZE} - 1)$\;
    }
    $\_\_syncthreads()$\;
    \If{$SMem[threadIdx.x] != 0$}{
      $\textnormal{atomicAdd}(\&VoxelHist[SMem[threadIdx.x] \gg 16], SMem[threadIdx.x] \& \textnormal{0XFFFF})$\;
    }
     
}

\vspace{0.2em}
\rule{\linewidth}{0.5pt}
\vspace{0.2em}

\tcp*[l]{CPU-side Memory Offset Computation}
\KwIn{VoxelHist}
\KwOut{AddrOffset}

$HeadAddr \gets 0$\;
\ForEach{VoxelIndex $\in$ VoxelNumber}{
    $cnt \gets (VoxelHist[VoxelIndex] == 0) \ ? \ 2 \ : \ VoxelHist[VoxelIndex] + 2$\;
    $VoxelHist[VoxelIndex] \gets HeadAddr$\;
    $HeadAddr \gets HeadAddr + cnt$\;
}
\textnormal{Memcpy}(AddrOffset, VoxelHist)\;
\end{algorithm}
Algorithm~\ref{alg1} illustrates the pseudocode for address offset generation.
The heterogeneous design is motivated by two considerations: (1) the voxel
occupancy histogram can be efficiently constructed in parallel on GPUs, while
the dilation offsets must explicitly distinguish between empty and non-empty
voxels that can be handled more flexibly on the CPU; (2) on embedded platforms,
unified memory architectures provide higher effective bandwidth and lower
coordination overhead than discrete GPUs. 
It worth to be noted that, since the cost of atomic operations on embedded GPUs
is significantly higher than discrete GPUs, constructing a histogram through the
global memory directly is not a good choice. Therefore, we propose a two-layer
memory structure based on a variant hash table strategy to reduce atomic
conflicts in global memory. Concretely, the points are first voxelized according
to their spatial coordinates (lines 2,3). We then derive hash keys from these
voxel indices and construct a hash table for each block in shared memory (lines
4-11). To avoid additional memory overhead that may reduce thread parallelism,
the hash key and value are compactly encoded into a single 32-bit entry, with
the upper 16 bits storing the key and the lower 16 bits storing the value. This
fused representation allows us to count the number of points while preserving
the original voxel index information. After the hash tables are filled, each
thread extracts the entries from non-empty hash slots and adds them to global
memory, where a global histogram over all voxels is subsequently constructed
(lines 12, 13).
Based on the histogram, the address offset of each voxel is then accumulated
according to the number of points it contains (lines 15-19). To save the memory
consumption, the offset values are updated in place directly on the
histogram. Furthermore, two additional segments are allocated for each voxel to
ensure correct parallel execution of the subsequent dilation step
(Figure~\ref{fig3}). Finally, the address offsets are copied into the device
memory region (line 20), since the offsets will be frequently accessed by
subsequent GPU kernels.
\begin{algorithm}[t]
\small
\DontPrintSemicolon
\SetAlgoNlRelativeSize{-1}
\caption{Voxelization under Dynamic Memory Assignment}
\label{alg2}

\KwIn{Target point $P_i$, AddrOffset}
\KwOut{VoxelMem}

$V_x, V_y, V_z \gets \textnormal{Quantize}(P_i.x, P_i.y, P_i.z)$\;
$VoxelIndex \gets \textnormal{Concat}(V_x, V_y, V_z)$\;
$VoxelAddr \gets AddrOffset[VoxelIndex]$\;

$lock \gets \textnormal{True}$\;
\While{$lock$}{
    \If{$\textnormal{atomicCAS}(VoxelMem[VoxelAddr], 0, 1) == 0$}{
        $count \gets VoxelMem[VoxelAddr + 1]$\;
        $VoxelMem[VoxelAddr + count + 2] \gets \textnormal{PointIndex}$\;
        $VoxelMem[VoxelAddr + 1] \gets count + 1$\;
        $\textnormal{atomicExch}(VoxelMem[VoxelAddr], 0)$\;
        $lock \gets \textnormal{False}$\;
    }
}
\end{algorithm}

\begin{algorithm}[t]
\small
\DontPrintSemicolon
\SetAlgoNlRelativeSize{-1}
\caption{Dilation under Dynamic Memory Assignment}
\label{alg3}

\KwIn{VoxelMem, AddrOffset}
\KwOut{VoxelMem (dilated voxel memory)}

\For{$iter \gets 0$ \textbf{to} DilationNumber}{
    $VoxelAddr \gets AddrOffset[VoxelIndex]$\;
    \If{$VoxelMem[VoxelAddr + 1] \neq 0$}{
        $DilationAddr \gets AddrOffset[VoxelIndex + k]$ \tcp*[r]{$k \in \{\pm 1, \pm 2^N, \pm 2^{2N}\}$}
        \If{$\textnormal{atomicCAS}(VoxelMem[DilationAddr], 0, 1) == 0$}{
            \If{$VoxelMem[DilationAddr + 1] == 0$}{
                $VoxelMem[DilationAddr + 1] \gets \textnormal{Mask}(VoxelIndex)$\;
            }
        }
    }
}
\end{algorithm}

\begin{algorithm}
\small
\DontPrintSemicolon
\SetAlgoNlRelativeSize{-1}
\caption{Fast Nearest Neighbor Search}
\label{alg4}
\KwIn{VoxelMem, AddrOffset, TargetSet, SourceSet}
\KwIn{Best\_Dist, Best\_Index}
  
$V_x, V_y, V_z \gets \textnormal{Scale}(S.x, S.y, S.z)$\;
$VoxelIndex \gets \textnormal{Concat}(V_x, V_y, V_z)$\;
$VoxelAddr \gets AddrOffset[VoxelIndex]$\;
$checkpoint \gets VoxelMem[VoxelAddr + 1]$\;
$IndexRelease \gets \textnormal{Unmask}(checkpoint)$\;

\If{$IndexRelease$ is \textnormal{DilatedVoxel}}{
    $VoxelAddr \gets AddrOffset[IndexRelease]$\;
    $count \gets VoxelMem[VoxelAddr + 1]$\;
}
\Else{
    $count \gets checkpoint$\;
}

\If{$count == 0$}{
    \ForEach{$P_i \in TargetSet$}{
        $Best\_Dist, Best\_Index \gets \textnormal{Globalsearch}(S, P_i)$\;
    }
}
\Else{
    \For{$i \gets 0$ \textbf{to} $count - 1$}{
        $P_i \gets TargetSet[VoxelMem[VoxelAddr + 1 + i]]$\;
        $Best\_Dist, Best\_Index \gets \textnormal{Localsearch}(S, P_i)$\;
    }
}
\end{algorithm}

\begin{figure*}[t]
\centering
\includegraphics[width=6.8in]{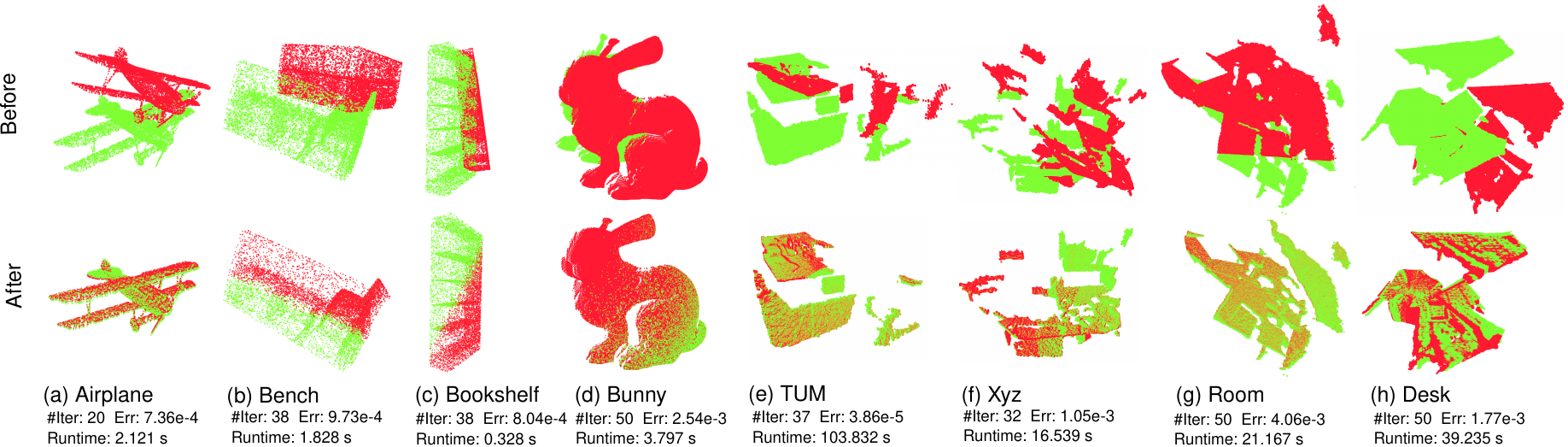}
\caption{Registration results of the dilation-based ICP applied to the source point cloud (red) and the target point cloud (green).}
\label{fig4}
\end{figure*}

Algorithms~\ref{alg2}--\ref{alg4} detail the voxelization, dilation, and nearest
neighbor search (NNS) procedures under the proposed dynamic memory assignment
strategy. They inherit the core computational design of VANICP, but employ a
different memory access strategy. VANICP achieves high throughput through a
monolithic memory policy, in which all point data are consolidated during the
voxelization stage into a single block, enabling dilation and NNS to be executed
by direct addressing. In contrast, the proposed method adopts a segmented
pointer-based assignment strategy, where points are accessed indirectly through
the index without being merged into a unified memory space (line 8 in
Algorithm~\ref{alg2}). This design substantially reduces the memory utilization
while preserving computational efficiency, as it eliminates the need for data
replication.  Figure~\ref{fig3} illustrates the overall memory assignment
strategy. The address offset, dilated voxel index, and points are stored in
three separate memory spaces while maintaining their logical dependencies
(highlighted in green and yellow). In contrast to VANICP’s static assignment
strategy, the proposed design exploits the address offset structure to achieve
dynamic and memory compact assignment (e.g. red-boxed area). To avoid race
conditions during the dilation stage, a {\em Thread Lock} is associated with
each voxel. During dilation, each voxel expands toward its six neighboring
directions. If a neighboring voxel is initially empty, the index of the root
voxel is bitwise-masked and retained until the NNS stage, after which it is
released (lines 3-7 in Algorithm~\ref{alg3}).
The NNS is then performed directly on the voxel index memory. The source points
are first voxelized and mapped to their corresponding voxel regions by an
address-offset lookup (lines 1-3 in Algorithm~\ref{alg4}). Each voxel entry is
subsequently examined by unmasking its state to determine whether it is
dilated. If not, a localized NNS is conducted using only the point indexes
within that voxel. Otherwise, the algorithm traces back to the associated root
voxel through its index linkage (lines 4-10). When the voxel remains empty, it
indicates that the boundary region of the space has not been fully filled,
requiring a fallback to a global search. Since such boundary cases occur
infrequently and involve only a small number of points, their overall impact on
runtime performance is negligible (lines 11-17).

\section{Experiments}
This section presents the experimental setup and a series of comparative
evaluations. Specifically, we benchmark performance across multiple datasets on
various platforms, including ARM, embedded GPU and FPGA.

\subsection{Experimental Setup}
\noindent\textbf{Platform} 
We employ the Jetson AGX Xavier embedded GPU as the primary experimental
platform. Built on the Volta architecture, it features 512 CUDA cores, providing
strong parallel compute capability for edge workloads. The module integrates 32
GB of LPDDR4x memory with a bandwidth of up to 137 GB/s and supports GPU boost
frequencies of approximately 1.37 GHz. Its unified global memory design allows
the CPU and GPU to access shared memory without explicit data-copy operations,
making the platform highly suitable for real-time and memory-constrained
embedded applications. Additionally, for comparison, we also implemented our
method on the Xilinx Zynq UltraScale+ MPSoC EV-series FPGA platform. It
integrates a heterogeneous architecture consisting of a Processing System (PS)
and Programmable Logic (PL). The PS includes four ARM Cortex-A53 cores operating
at up to 1.2 GHz with a two-level cache hierarchy, while the PL offers 88 K
lookup tables, 176 K flip-flops, and 4.5 MB of on-chip BRAM for custom hardware
acceleration.
\noindent\textbf{Dataset} 
We evaluated our method on three point cloud datasets of different scales: {\em
  ModelNet}~\cite{wu20153d} with an average of 10k points, the {\em Stanford
  Bunny}~\cite{turk1994zippered} with approximately 40k points, and the {\em
  TUM} RGB-D dataset~\cite{sturm2012benchmark} containing over 200k
points. These datasets are widely used in point cloud processing research and
offer strong benchmarking value for performance assessment. All point clouds
were normalized to fit within a unit sphere, and random perturbations were
applied following the same protocol as the original
VANICP~\cite{chang2025accelerating}.
\noindent\textbf{Parameter} 
In dilation-based ICP registration, four parameters play essential roles: the
voxelization scale $N$, the number of dilation layers $L$, the maximum iteration
count and the {\em RMSE} difference, both of which serve as convergence
thresholds. The number of voxels along each dimension is $2^N$. VANICP is highly
sensitive to $N$ because it directly determines voxel-wise memory assignment. In
contrast, our method is only weakly influenced by this parameter; therefore, we
fix $N=4$ in our experiments. The parameter $L$ is inversely related to $N$: a
smaller voxel size requires more dilation steps to adequately cover the entire
space, vice versa. Accordingly, we set $L=10$. The maximum iteration count and
the {\em RMSE} difference thresholds are set to 50 and
\begin{math}
1\times{e^{-5}},
\end{math}
respectively, to ensure stable convergence across all tested datasets.

\begin{table*}[h]
    \centering
    \caption{Memory and Energy Consumption Analysis}
    \label{tb1}
    \resizebox{1\textwidth}{!}{
    \begin{tabular}{l|c|c|c|c|c|c|c|c|c|c|c|c|c}
        \hline
        \multirow{2}{*}{Dataset} & \multicolumn{2}{c|}{GPU Mem. (MB)} & \multirow{2}{*}{Saving} & \multicolumn{3}{c|}{Runtime}&\multicolumn{4}{c|}{Power(w)$^3$}&\multicolumn{2}{c|}{EC(J/F)$^4$}&\multirow{2}{*}{Saving}\\
\cline{2-3}\cline{5-13}
        &\textbf{Ours$^1$}&VANICP&&MOC$^2$(ms)&\textbf{Ours Total(s)}&VANICP Total(s)&SYS\_GPU&SYS\_SOC&SYS\_CPU&SYS\_DDR&\textbf{Ours}&VANICP&\\
        \hline
        AirPlane (10k)&0.135&10&\textbf{98.6\%}&0.251&2.12&2.089&0.458&1.07&0.458&0.152&4.7&6.23&\textbf{24.5\%}\\
        Bench (10k)&0.135&6.25&\textbf{97.8\%}&0.207&1.83&2.156&0.763&2.29&2.28&0.458&4.94&5.9&\textbf{16.4\%}\\
        Bookshelf (10k)&0.135&5&\textbf{97.3\%}&0.328&2.3&2.07&0.763&2.138&1.832&0.458&6.122&5.84&-4.1\%\\
        Bunny (40k)&0.307&75&\textbf{99.6\%}&0.399&3.8&2.356&7.771&3.2&0.913&0.913&11.48&10.417&-10.2\%\\
        TUM (226k)&1.39&1508&\textbf{99.9\%}&1.29&103.8&156.01&9.74&3.65&0.608&1.065&330.7&471.74&\textbf{29.8\%}\\
        Xyz (229k)&1.39&312&\textbf{99.5\%}&1.27&16.54&28.477&9.29&3.351&0.913&0.761&50.675&84.94&\textbf{40.34\%}\\
        Room (239k)&1.39&381&\textbf{99.6\%}&1.34&21.17&49.85&9.14&3.352&0.609&0.761&64.01&154.73&\textbf{58.63\%}\\
        Desk (250k)&1.53&450&\textbf{99.6\%}&1.42&39.2&73.42&9.9&3.351&0.456&0.761&118.64&224.95&\textbf{47.2\%}\\ 
        \hline
    \end{tabular}
    }
     \parbox{1\linewidth}{\footnotesize $^1$ Static memory utilization for the
       \textbf{Histogram} and \textbf{Address offsets}: 0.03125 MB + dynamic
       memory utilization of the \textbf{Dilated Voxel Index}. $^2$: Memory offset
       computation. $^3$: Power consumption of GPU, CPU, SoC and memory. $^4$:
       Energy consumption of the entire system per frame.}
\end{table*}

\subsection{Experimental Results}
We selected eight representative models from three datasets, covering a broad
range of point counts, sparsity levels, and initial spatial
poses. Figure~\ref{fig4} shows the registration results obtained using our
proposed strategy. As the model size increases, both the number of iterations
and registration time increase accordingly. Although the registration time and
accuracy exhibit minor variations due to differences in initial poses, our
strategy demonstrates strong robustness. In all cases, the source point cloud
(red) successfully converges to the target (green).
Table~\ref{tb1} summarizes the memory and energy consumption of different models
on the Jetson AGX Xavier GPU, including GPU memory utilization, runtime, and
power consumption. For comparison, we also report the results obtained by
VANICP. Our strategy comprises two memory components: a static footprint for the
Histogram and Address Offsets (0.03125MB), and a dynamic footprint for the
Dilated Voxel Index. Even for large-scale point clouds such as {\em Desk} with
250k points, the total memory required for dilation is only 1.53MB. In contrast,
VANICP employs a static memory allocation strategy without voxel-wise
adaptivity, necessitating a large pre-allocated memory space to guarantee
successful dilation. For large and spatially non-uniform point clouds (e.g.,
{\em TUM}), the memory size even exceeds 1.5GB. Consequently, our dynamic memory
assignment strategy reduces memory utilization by more than 97\%.
Notably, our dynamic memory assignment strategy does not incur additional
computational overhead, as the memory-offset computation (MOC) consistently
remains below 2ms. For small point clouds (fewer than 40k points), the runtime
of our strategy is comparable to that of VANICP. For large-scale point clouds,
the registration time of VANICP even increases, since points that exceed the
voxel capacity force the algorithm to fall back to global registration.
The {\em Power} and {\em EC} columns summarize the power consumption of each
model on embedded GPU, including the GPU, CPU, SoC, and DDR memory, as well as
the energy consumed per frame. Power consumption increases proportionally with
the size of the point cloud model. Overall, our system exhibits slightly higher
instantaneous power consumption than VANICP across most models. However, since
our strategy achieves shorter runtimes, its overall energy efficiency surpasses
that of VANICP for all models except {\em Bookshelf} and {\em Bunny}, resulting
in energy savings between 16.4\% and 58.63\%.

\begin{table}
    \centering
    \caption{Performance comparison across multiple platforms}
    \label{tb2}
    \resizebox{0.5\textwidth}{!}{
    \begin{tabular}{c|l|c|c|c|r}
        \hline
        Dataset & Methods & Hardware$^1$ & Runtime(s) & Speedup & Error \\
        \hline
        \multirow{5}{*}{ModelNet} %% & KD-PCL\cite{rusu20113d} & CPU & 1.86 & 1x & 1e-6 \\
        %% & Open3D\cite{zhou2018open3d} & CPU & 0.15 & 12.4x & 5.5e-3 \\
        %% & PQT\cite{wieschollek2016efficient} & CPU & 1.29 & 1.4x & 8.44e-4 \\
        & FRICP\cite{zhang2021fast} & ARM & 0.17 & 1x & 2.08e-6 \\
        & Brute & EmGPU  & 2.5 &0.07x &   \textbf{5.08e-7}\\ 
        & KD-tree\cite{project4} & EmGPU & 5.35 & 0.03x & 9.9e-5 \\
         (10k)& VANICP\cite{chang2025accelerating} & EmGPU & 2.089 & 0.08x & 5.4e-4 \\
\cline{2-6}
        & \multirow{2}{*}{Ours} & FPGA & \textbf{0.12} & \textbf{1.4x} & 1.38e-3 \\
        &  & EmGPU & 2.121 & 0.08x & 7.36e-4 \\
        %&  & GPU & 0.16 & 1.06x & 6.3e-4\\
        \hline
        \multirow{5}{*}{Bunny} %% & KD-PCL\cite{rusu20113d} & CPU & 7.2 & 1x & 8e-3 \\
        %% & Open3D\cite{zhou2018open3d} & CPU & 0.67 & 10.7x & 7.1e-3 \\
        & FRICP\cite{zhang2021fast} & ARM & 15.2 & 1x & 3.05e-3\\
        & KD-tree\cite{project4} & EmGPU & 8.1 & 1.87x  & \textbf{1.28e-4} \\
        & Brute & EmGPU & 5.01 & 3x& 8e-3\\
        (40k)& VANICP\cite{chang2025accelerating} & EmGPU &3.356 & 4.5x & 9e-3 \\
\cline{2-6}
        & \multirow{2}{*}{Ours} & FPGA & \textbf{0.6} & \textbf{25x} & 1.47e-2 \\
        & & EmGPU & 3.793 & 4x & 2.54e-3 \\
        %& & GPU & \textbf{0.3} & \textbf{24x} & \textbf{2.02e-5}\\
        \hline
    \end{tabular}
    }
    \parbox{1\linewidth}{\footnotesize $^1$ \textbf{ARM}: Cortex-A53. \textbf{EmGPU}: Jetson AGX Xavier. \textbf{FPGA}: Zynq UltraScale+ MPSoC EV.}
\end{table}

Table~\ref{tb2} presents the results of comparing existing methods across
multiple mobile platforms. FRICP demonstrates excellent efficiency on the
small-scale {\em ModelNet} dataset, significantly outperforming embedded
GPU–based approaches. However, its performance degrades as the dataset size
increases. Among the embedded GPU methods, our strategy achieves performance
comparable to VANICP and attains nearly twice the speed of the remaining
methods. It is also worth noting that we include experimental results on an FPGA
platform, which achieves the highest speed among all methods. However, its
accuracy is lower due to the use of integer-based processing.

\section{Related Work}
In recent years, numerous studies have focused on accelerating the processing of
the ICP algorithm through various approaches to promote the practical
applications of point cloud technology.
Zhang et al. treated the classic ICP algorithm as a Maximization-Minimization
(MM) algorithm~\cite{zhang2021fast}, using Anderson acceleration to speed up the
algorithm's convergence. This approach reduces the number of iterations to
improve the algorithm's processing speed and minimizes the error parameters
based on the Welsch function introduced in the method.
Li et al. used a K-D tree \cite{li2016tree} data structure to organize scattered
point cloud data, improving the efficiency of nearest neighbor search in the
algorithm. However, the K-D tree-based search process requires a large number of
non-continuous memory accesses, placing high demands on memory bandwidth.
In \cite{kosuge2020soc}, the authors proposed a hierarchical graph-based KNN
search method and the corresponding hardware sorting network to accelerate the
ICP algorithm. This method organizes data points into a graph structure with
multiple hierarchical levels, achieving faster and more efficient searches by
progressively narrowing the search space, which offers a significant advantage
over traditional K-D tree search methods. Li, Zheng, and Xiao, based on an
approximate K-D tree structure, proposed an efficient parallel sorting circuit
\cite{li2022knn} to accelerate the sorting of large amounts of data during tree
construction, while simultaneously computing the Euclidean distances of all
points in the subspace. However, this approximate K-D tree-based method is prone
to finding incorrect nearest neighbor target points, which leads to the ICP
algorithm requiring more iterations.
In \cite{wang2024fpga}, Wang et al. proposed an improved Local Sensitive Hashing
(LSH) method, which can assign similar points to the same hash bucket
(low-dimensional space), thereby reducing the search space and accelerating kNN
search speed. This method requires the selection of an appropriate hash function
to ensure that similar points are assigned to the same bucket.
In \cite{feng2022fast}, the authors designed multiple units to store point cloud
data. These units perform parallel nearest neighbor searches, and the sum of the
absolute differences in the coordinates of the source and target points in three
dimensions replaces the traditional Euclidean distance calculation. This method
of using the sum of coordinate differences instead of Euclidean distance
calculation may lead to larger errors and increase the number of iterations of
the algorithm.

\section{Conclusion}
In this work, we introduced an embedded GPU-oriented ICP accelerator for point
cloud registration. The proposed strategy employs a dynamic memory assignment
strategy that adapts to the point distribution within each voxel block,
substantially reducing memory consumption. Moreover, by integrating a voxel-wise
dilation scheme, the accelerator converts global nearest-neighbor queries into
localized searches, effectively lowering computational demands without
compromising registration accuracy.

Looking ahead, we plan to enhance both the operational frequency and the
precision of the accelerator, and to explore its applicability across a wider
range of mobile computing platforms.

\bibliographystyle{unsrt}  
\bibliography{refs}

\end{document}